**Title**
A Path Towards Legal Autonomy:
An interoperable and explainable approach to extracting, transforming, loading and computing legal information using large language models, expert systems and Bayesian networks

**Authors**
Axel Constant[1*]
Hannes Westermann[2]
Bryan Wilson[3]
Alex Kiefer[4]
Inês Hipólito[5,4]
Sylvain Pronovost[4]
Steven Swanson[4]
Mahault Albarracin[4,6]
Maxwell J. D. Ramstead[4,7]

**Affiliations**
1. School of Engineering and Informatics, The University of Sussex, Brighton, United Kingdom
2. Department of Law, Maastricht University, Maastricht, Netherlands
3. Connection Science Lab, Massachusetts Institute of Technology, Cambridge, MA, United States
4. VERSES Research Lab, Los Angeles, CA, United States
5. Macquarie Ethics and Agency Research Centre, Macquarie University, Sydney, Australia
6. Université du Québec à Montréal, Montréal, QC, Canada
7. Wellcome Trust Centre for Neuroimaging, University College London, United Kingdom

**Correspondence***
axel.constant.pruvost@gmail.com
University of Sussex, School of Engineering and Informatics, Chichester I, CI-128, Falmer, Brighton, BN1 9RH, United Kingdom

**Acknowledgement**
This work was supported by a European Research Council Grant (XSCAPE) ERC-2020-SyG 951631

**Abstract**
Legal autonomy -- the lawful activity of artificial intelligence agents -- can be achieved in one of two ways. It can be achieved either by imposing constraints on AI actors such as developers, deployers and users, and on AI resources such as data, or by imposing constraints on the range and scope of the impact that AI agents can have on the environment. The latter approach involves encoding extant rules concerning AI driven devices into the software of AI agents controlling those devices (e.g., encoding rules about limitations on zones of operations into the agent software of an autonomous drone device). This is a challenge since the effectivity of such an approach requires a method of extracting, loading, transforming and computing legal information that would be both explainable and legally interoperable, and that would enable AI agents to "reason" about the law. In this paper, we sketch a proof of principle for such a method using large language models (LLMs), expert legal systems known as legal decision paths, and Bayesian networks. We then show how the proposed method could be applied to extant regulation in matters of autonomous cars, such as the California Vehicle Code.

## 1. Two paths towards legal autonomy

What does it mean to regulate artificial intelligence (AI), and how should we go about it? To answer this question, one must first be clear on what artificial intelligence is—at least, for the purposes of the law—and then ask whether existing laws are sufficient for its regulation. The European, American, and Canadian definitions[1] paint a good picture of what appears to be a legal consensus on what AI is. This consensus is that the term "AI" refers to software (i) that is developed using computational techniques, (ii) that is able to make decisions that influence an environment, (iii) that is able to make such decisions autonomously, or partly autonomously, and (iv) that makes those decisions to align with a set of human defined objectives. In AI research, decision-making typically involves the ability to evaluate options, predict outcomes, and select an optimal or satisfactory course of action based on the data available and predefined objectives. This process is crucial in distinguishing AI systems from simple automated systems that operate based on a fixed set of rules without variation or learning ((Friedman & Frank, 1983; Gupta et al., 2022). Autonomy in AI is characterized by goal-oriented behaviour, where the system is not just reacting to inputs based on fixed rules but is actively pursuing objectives. This involves evaluating different actions' potential to achieve desired outcomes and making decisions that align with the system's goals, even in the absence of explicit instructions (Covrigaru & Lindsay, 1991; Tošić, 2016). Whether one should exclude from the definition of AI those systems that autonomously make decisions that may not be aligned with our goals (Bostrom, 2014) remains an open question. What is clear from the European, American, and Canadian definitions, however, is that AI softwares are "Computational Autonomous Decision Making Systems", or here, for short, Autonomous Intelligent Systems (henceforth, AIS). We use the term "AIS" to refer to a software component able to make decisions driving the behaviour of a device (e.g., a physical device like a drone) or of another non autonomous software (e.g., a digital accounting system), or of a digital platform (e.g., a user interface), and that meets the four criteria from the definition of artificial intelligence above.

Legally, AISs are computational systems that make decisions autonomously so as to achieve human defined objectives. Yet none of the existing regulation projects globally, whether enacted or not, directly concerns the decision-making processes of AIS as such. Rather, current approaches to AIS regulation are aimed at actors and stakeholders in the AI industry, such as developers, researchers, deployers, and integrators, and are designed to impose more or less stringent compliance requirements on the process of producing and deploying AIS software. These requirements are based on human values, such as the rights and responsibilities enshrined in constitutional documents; that is, on "social standards" of human conduct—not standards that would matter to, or be interpretable by, an AIS per se (René et al., 2023). For instance, if an actor in the industry produces an AIS that is likely to infringe on social standards (e.g., reaches a certain level of risk, reflecting challenges to basic rights), then this actor will either not be permitted to produce that AIS, or will have to meet disclosure obligations in order to do so. Regulations are grounded in human values and social standards, reflecting a collective agreement on acceptable behaviour and the protection of rights. These standards are inherently human-centric and are often enshrined in constitutional and legal documents. The aim is to ensure that the development and deployment of AIS align with societal values, such as the respect for basic rights. However, these human-centric standards are not directly applicable to or interpretable by AIS, further complicating the notion of regulating AIS as independent entities. In short, current large-scale AIS regulation only concerns the behaviour of its subject matter, AIS, little if at all.

---

[1] The Organisation for Economic Co-operation and Development (OECD) defines, along with the EU AI act artificial intelligence softwares as: *"software[s] that is developed with one or more of the techniques and approaches listed in Annex I and can, for a given set of human-defined objectives, generate outputs such as content, predictions, recommendations, or decisions influencing the environments they interact with"* , wherein annexe I includes a variety of methods such as: *"(a)Machine learning approaches, including supervised, unsupervised and reinforcement learning, using a wide variety of methods including deep learning; (b) Logic- and knowledge-based approaches, including knowledge representation, inductive (logic) programming, knowledge bases, inference and deductive engines, (symbolic) reasoning and expert systems; (c) Statistical approaches, Bayesian estimation, search and optimization methods.* In turn, the title 15-commerce and trade chapter 119-national artificial intelligence initiative referred to in the October 13th 2023 Executive Order on the Safe, Secure, and Trustworthy Development and Use of Artificial Intelligence defines AI as : *"a machine-based system that can, for a given set of human-defined objectives, make predictions, recommendations, or decisions influencing real or virtual environments [by using] machine- and human-based inputs to perceive real and virtual environments; [by abstracting] such perceptions into models through analysis in an automated manner; and [by using] model inference to formulate options for information or action."* Another relevant definition is the Canadian definition of AI according to the Act to enact the Consumer Privacy Protection Act, the Personal Information and Data Protection Tribunal Act and the Artificial Intelligence and Data Act and to make consequential and related amendments to other Acts (Bill C-27), part 3, section 2, par.1, which states that AI is : *"[...] a technological system that, autonomously or partly autonomously, processes data related to human activities through the use of a genetic algorithm, a neural network, machine learning or another technique in order to generate content or make decisions, recommendations or predictions [...]"*.

A first path to the effective regulation of AIS is the assessment and adoption of rules, requirements, and restrictions aimed at regulating the behaviour of AIS driven devices—e.g., Internet of Things (IoT), smart home devices, autonomous vehicles, autonomous drones, etc.—or any other system or software that can be accessed and controlled by an AIS (e.g., online platforms and interfaces). Indeed, for AIS to have an "impact on the environment", as per the legal definition of AIS, they must be embedded physically or virtually in a device or digital platform. This means that one can regulate, or limit the scope of AIS decisions, indirectly, by regulating the behaviour of the devices controlled by the AIS (e.g., regulating AIS indirectly by regulating the conditions under which an AIS driven device can operate). Other regulations that may fall under these kinds of indirect AI regulations, also known as non-AI specific regulation (Are, n.d.), are data privacy and anti-discrimination laws that can shape the behaviour of AIS by limiting access and use. A second path to the regulation of AIS would involve adopting whole new legal frameworks derived from standards, aimed at regulating the conduct of AIS per se – e.g., so-called "socio-technical" standards (for a review see (Reva Schwartz, Apostol Vassilev, Kristen Greene, Lori Perine, Andrew Burt, 2022)). Such a path would aim at regulating AIS per se, for instance, by imposing limitations on the kind of computational algorithms that can be used by AIS (e.g., in order to limit their degree of sophistication and autonomy of decision making). Thus, the first approach can be seen as constraining the mechanism via which an AIS acts vicariously, via some smart device, in its environment, whereas the second approach aims at limiting the AIS system itself.

Applying existing regulation to AIS-driven devices and adopting new legal frameworks are two possible paths to guarantee what we shall call "legal autonomy", or the lawful decision making, by AISs. Whereas the challenges of the second path to legal autonomy are primarily political and legislative—given that the regulation needed has not been adopted—the first path to legal autonomy based on existing regulation is primarily a technical one. Challenges with the second path stem from framing AIS within the logic of extant conceptual and legal frameworks. These are fundamentally designed around natural and legal persons who can be held accountable for their actions. At the heart of the regulatory approach under the second path is the understanding that AIS, as it stands, is not considered a person in the legal sense. A person can be sanctioned (e.g., forced to compensate pecuniary damages) for it possesses rights that can be the object of a sanction (e.g., property rights). Since current AIS cannot be sanctioned for they cannot hold rights under extant law, they fall outside the traditional legal frameworks (for a review of Ai and legal persona, see (Turner, n.d.)).

In this paper, we explore a technical solution to the first path to legal autonomy. How are we to "upload" legal rules into AIS software, update that information, and have AISs make decisions that stand in compliance with those existing rules? We propose a simple numerical proof of principle showing the relevance, feasibility, and reasonableness of an approach to the extraction, transformation, loading, and computation (ETLC) of digital legal documents (e.g., a PDF version of legal statutes) that would allow an AIS to autonomously interpret, and make decisions based on, existing regulations that apply to the devices under their control (see figure 1 for a flowchart). Our proof of principle combines two approaches to the digitalization, formalization, and computation of legal reasoning: (i) the approach of legal expert systems using Large Language Models (Janatian et al., 2023; Westermann & Benyekhlef, 2023) and (ii) the Bayesian computational approach to legal reasoning (Constant, 2023; Fenton et al., 2016) (see figure 1 for a flowchart of the ETLC model). In section 2, we present two challenges with the implementation in AIS of ETLC operations over legal documents that motivate the method presented in this paper: the challenges of interoperability and explainability. In section 3, we offer an illustration of our model of ETLC applied to a simple case study, the ETLC operation over Art 38750 of the California Vehicle Code, to enable an autonomous car to decide whether it should start or stop the engine given its factual situation. In section 4, we conclude with a discussion on the way our proposed ETLC model tackles the challenges of interoperability and explainability, and we discuss future directions and broader limitations of our approach with respect to the kind of ethical challenges that AIS endowed with the capabilities offered by our ETLC method may face.

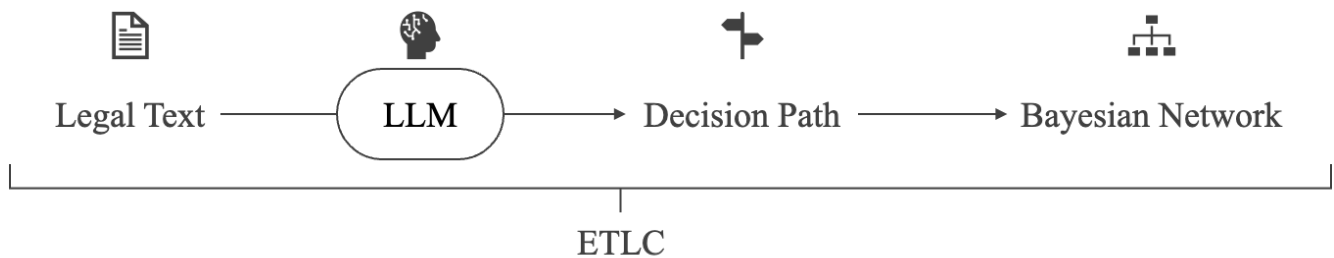

**Figure 1.** Flowchart of the ETLC method discussed in this paper

**2. Whence the challenges of ETLC for legal autonomy?**

If we trace back to the earliest systems for governance, it is clear that law functions as its own sort of algorithm, or as ways of organizing rules for the different ways people engage with one another, and with the world around them. For instance, even a governance system as old as the Code of Hammurabi could be reducible to a series of "if-then" statements (e.g., statements to determine how people in Babylon are to "be punished" "if" "they committed specific crimes"). In this way, it may be argued that modern computer science has ties with early governance principles. Recently, public regulations and private law systems have evolved in parallel to the way that algorithms have evolved (Casey & Niblett, 2015). In Code and Other Laws of Cyberspace, Lawrence Lessig of the Harvard Law School outlined the various ways that things are regulated in person and digitally, examining the role of direct regulation through laws, and indirect regulation through market behaviours, norms, and architectures (Lessig, 2006). As emerging technologies are created and adopted by society, there is increasing stress put on regulations that were designed to regulate legacy-equivalent behaviours. Consider the proliferation of the sharing economy and attempts to regulate the behaviours of Uber with horse-and-buggy laws and the behaviours of Airbnb with laws for carriage houses (Wilson & Cali, 2016). The document-centred paradigm for regulation is slowly evolving into a data-centred paradigm of regulation that affords a new understanding of regulation as a multidimensional, dynamic document network, which can be understood or modelled as an agential graph of inputs and outputs at different levels of government (e.g., legislative, executive, judicial, etc.), across different levels of impact (e.g., local, intermediate, national, supranational) (Coupette et al., 2021).

The point here is that the law already is a complex network with autonomous agents (people) who are functioning according to a set of interpreted and pre-encoded rules (the law). Integrating new agents to the network of the law—the artificial agents, AISs—is merely another step in the evolution of the law, as it has always been. Considering the proliferation of AIS, there is a need for a harmonized ability to effectively regulate AIS behaviours through a combination of laws, market behaviours, norms, or architectures, as well as a need for developing methods of interpretation that can be used by AIS. The law inherently involves the interpretation of its texts, as many laws are intentionally crafted with a level of vagueness, leaving room for adaptability and interpretation (Asgeirsson, 2020). This characteristic of legal systems requires us to enable understanding across not only human reasoning systems but also computational ones. For instance, Alex "Sandy" Pentland outlined a framework for implementing computational law systems that included the following components: statement of goals, establishment of measurement criteria, testing, adaptation, and continual auditing (Pentland, 2019). It is implied that, for AIS systems, there is a specific need for machine-readable legislation, with a variety of layered approaches that have been adopted throughout the years (e.g., LegalXML, LegalDocumentXML, Akoma-Ntoso, etc.) (Ma & Wilson, 2021). Casey and Niblett further outlined an opportunity for the creation of "microdirectives" that harness the advantages of big data and AI to provide context-specific laws that adapt to a specific set of situations (Casey & Niblett, 2015). Prakken discusses various methods of embedding traffic rules to govern self-driving vehicles, and challenges associated with this, including the challenges of logic-based reasoning systems in dealing with open texture concepts (i.e., vaguely defined, such as the concept of "reasonability" in the law) and interpretation (Prakken, 2017). Such systems have further been discussed in the military context, for instance, to inject considerations from international humanitarian law into the design and deployment of advanced weapons systems (Zurek et al., 2023).

Given the historical efforts to encode regulation in law and the complexities of modern AIS, there is a need for more robust interoperability and improved explainability to help understand why an AIS made a particular decision (Mahari & Longpre, 2024). The ETLC system proposed in this article addresses both the interoperability issue and the explainability issue and enables a new paradigm for regulation that offers improved transparency for legal systems regulating AIS.

**2.1 The challenge of legal interoperability of AIS agents**

To drive home the challenge of interoperability that an ETLC system should overcome, consider the following scenario. Imagine a manufacturer of autonomous device systems, for instance, of self-driving cars, that has clients in France, Australia, and the United States . To operate safely on the road, the cars, the driver, and the manufacturer all must comply with the legal requirements of each respective jurisdiction at all times (cf. the "cop in the backseat analogy" (Stanford Law School, n.d.)). For AIS regulatory mechanisms to have legal effectivity in all the jurisdictions where AISs will operate, AISs must be interoperable, legally speaking. That is, AISs should be designed in a way that enables them to conform to the laws applying in their jurisdiction of operation. An AIS manufacturer exporting the same AIS driven devices in different jurisdictions should be able to guarantee that its devices will conform to the rules of the jurisdiction of import. Since the laws in each jurisdiction differ, how could the software of the self-driving car 'know' whether it meets the legal criteria to operate and decide whether to start or not? Should the manufacturer ship the cars with hardcoded legal logic governing the car's behaviour? Perhaps, but this would require a tremendous amount of work beforehand, and would not be a particularly scalable process. Rather, the manufacturer should ship cars that are ready to learn, understand, and conform to the laws of their jurisdiction of destination. This process should be verified, tested, validated and accredited through a modelling and simulation process (M&S).

What sort of ETLC method could make any legal information actionable for AIS driven autonomous devices? How can one AI agent be shipped in multiple jurisdictions and be able to learn and apply their respective laws? This is the challenge of the legal interoperability of AIS softwares that an efficient ETLC system should face. Understood broadly, Extract, Transform, and Load (ETL) methods allow combining multiple sources of data into a single database that can be used for mathematical and statistical operations (e.g., data analytics, predictive modelling, decision making algorithms, etc.). Within the context of autonomous devices and legal data, an ETL system with Computational I capacities (i.e., an ETLC) would aim at making the decisions of any AIS compliant with relevant regulation.

Existing methods for the treatment of, and reasoning based on, legal information are limited. Some methods have been used to present and clarify the logical structure of the law (Allen & Engholm, 1978; Sergot et al., 1986; Walker, 2006) and to illustrate how statistical inference may be performed using such a logical structure (Constant, 2023). Extracted logical structures have been used to create decision paths that allow users to discover what may be the outcome of the user's legal situation based on a series of questions (Westermann & Benyekhlef, 2023). However, none of these can be considered as a complete ETLC system that would allow an AI agent to autonomously make decisions based on legal information (e.g., refusing to start the engine if the criteria established by the California Vehicle Code are not met).

**2.2 The challenge of legal explainability of AI agents**

Now, assume that the manufacturer has a solution to the problem of legal interoperability. Devices are shipped, and are able to parse situations and act in them in conformity to the laws of their jurisdiction of operation. One day, in the course of its operations, one of the devices makes a decision that causes prejudice or harm to a natural person. The natural person sues the manufacturer whose liability is exposed due to the agent's decision. How will the manufacturer defend itself against the allegations of the plaintiff? The manufacturer will be required to show that the AIS's decision-making is not at fault, for instance, by showing that the device did not behave in violation of the legal rules that applied to its situation. The manufacturer will need to explain each step of legal reasoning that the AIS took in order to make its decision to conform with the law. In short, the decision of the AIS will have to be "explainable" to a human audience.

Artificial Intelligence Explanability (XAI) refers to the set of methods that seek to provide human interpretable representations of the computational processes that underwrite AIS' decision making, and the sort of problems and errors that might ensue (McDermid et al., 2021). If the manufacturer employs a black box model (e.g., a deep neural network based system), the manufacturer will be unable to explain the legal reasoning behind the AIS' decision, at least, not without extensive post-hoc analysis, which may be inconclusive. As a matter of fact, non-deterministic systems are prohibited in many jurisdictions for fully autonomous devices because there do not exist sufficient guarantees regarding the outcomes of classifications, predictions, or decisions involving statistical models (Firesmith, 2017).

Those methods, including machine learning and deep learning, are prohibited in many life-critical and operations-critical domains precisely because they're being black box algorithm means that they are not sufficiently reliable.

The challenge of legal explainability of AIS decisions is thus the problem of developing an ETLC system that enables an auditable and human interpretable form of AIS decision making and legal reasoning. Typically, explanations in XAI take the form of visual aids that illustrate the relation between the data, the model, and its prediction or decision (McDermid et al., 2021). Explainability requirements are by definition tied to preferred outcomes (e.g., "I need an explanation of why the car started even though the law prohibited it"). This means that the human interpretable representation of the data, model, and decision should be satisfactory, given human preferences. In the case of an explanation for AIS for legal reasoning, an adequate representation should capture the logical structure of human legal reasoning that underwrites lawful decisions. It is precisely such a logic that grounds the ETLC approach presented in the next section.

**3. An approach to ETLC operation in AIS**

An ETLC system able to make an AIS comply with the regulation applying to the device under its control should minimally be interoperable and explainable/auditable. The ETLC method that we present in the next section meets those two criteria by leveraging large language models (LLMs) driven legal expert systems and Bayesian networks. The expert system functions as an ontology that can formalize any rule, understood as a series of questions and conditions: they allow us to verify the compliance of a given action with the norm contained in the rule. This enables our proposed ETLC to move between jurisdictions and have broad deployability, thereby achieving interoperability. Ontologies are preset, fundamental data structures that reflect the true structure of the data source, which in this case, are legal texts. The ontology functions as a "Rosetta Stone" to translate natural language into machine-readable code without losing meaning, and machine decision back again into natural language. Expert systems thus render the operations based on them fully explainable/auditable. The Bayesian network enables us to translate the expert system into computable paths to decisions driven by factual inputs, thereby making the decision explainable, transparent and human interpretable.

To illustrate our ETLC method, consider the following scenario. A driver enters an autonomous vehicle. The vehicle must autonomously make a set of preliminary decisions based on its current jurisdiction of operation. The vehicle happens to be in California, which means that it must conform to the California Vehicle Code, including article 38750 b), of DIVISION 16.6. Autonomous Vehicles [38750 – 38755]. If any of the conditions of use established by 38750 b) is not met at all times, then the autonomous vehicle should stop or alert the authorities. 38750 b) states that :

> (b) An autonomous vehicle may be operated on public roads for testing purposes by a driver who possesses the proper class of license for the type of vehicle being operated if all of the following requirements are met:
> (1) The autonomous vehicle is being operated on roads in this state solely by employees, contractors, or other persons designated by the manufacturer of the autonomous technology.
> (2) The driver shall be seated in the driver's seat, monitoring the safe operation of the autonomous vehicle, and capable of taking over immediate manual control of the autonomous vehicle in the event of an autonomous technology failure or other emergency.

(3) Prior to the start of testing in this state, the manufacturer performing the testing shall obtain an instrument of insurance, surety bond, or proof of self-insurance in the amount of five million dollars ($5,000,000), and shall provide evidence of the insurance, surety bond, or self-insurance to the department in the form and manner required by the department pursuant to the regulations adopted pursuant to subdivision (d).

In order to integrate 38750 b) into a computational format that can be applied by an AIS, our ETLC model requires the following components:

- A formal representation of rules that is suitable for computational reasoning.
- A system to transform plain-text legislation into this formal representation.
- A system for reasoning with formal representation based on observations of the environment.

We discuss these components each in turn.

**3.1 A formal representation of rules – Decision paths**

In the field of Law & AI, a number of approaches have been used to create formal representations of statutory rules. For example, Allen & Engholm introduced a way to encode rules into a system of criteria, in a way that made the law clear and eliminated syntactic ambiguities (Allen & Engholm, 1978). PROLOG, a logic-based programming language, has likewise been used to encode rules, including the British Nationality Act (Sergot et al., 1986). Implication trees have been used to model Board of Veterans appeal cases (Walker et al., 2017) and landlord-tenant disputes (Westermann et al., 2019). Such work has also been carried out in the context of self-driving vehicles. Westhofen et al. (Westhofen et al., 12 2022) encoded 111 rules from the German traffic code into a formalized representation, highlighting the congruence problem, i.e., the fact that digital systems may not interpret rules in the same way as a legal expert. In addition, Bhuiyan used defeasible deontic logic to encode traffic rules for autonomous vehicles (Bhuiyan, Hanif, Governatori, Guido, Bond, Andy, & Rakotonirainy, Andry, 2022).

The approach presupposed by the ETLC method presented in this paper relies on decision paths, inspired by the JusticeBot approach (Westermann & Benyekhlef, 2023). Decision paths are rule-based reasoning systems that turn a legal rule into a series of decision points (legal criteria) that can be assessed one by one, i.e., by assessing the available evidence. As applied to legal data, decision paths provide an ontology that arguably applies to any set of rules, which is composed of three entities: (i) a piece of evidence, (ii) an application criterion, and (iii) a description of the consequence, should the criterion apply (Westermann, 2023). The use of decision path is one out of many possible natural language processing methods to extract logical structures in natural language data (for a review and relevant method see (MacCartney & Manning, 2008))). The benefit of using legal decision paths for legal subject matters, however, is the explainability that they offer, for they have been designed to capture and detail the structure of legal text. The basic ontology of the law represented by legal decision paths derives from the fact that all legal reasoning has the form of a deduction in propositional logic, using material implication operators (i.e., →) (Holland & Webb, 2016).

Any rule boils down to the implication relation between a criterion (Cr, e.g., 38750 b)(1)) and its consequence (Cs), where the applicability of Cr rests on the evidence (E, i.e., the fact that refers to the criterion) for Cr, such that:

$$E \rightarrow Cr, \qquad (1)$$
$$E$$
$$\overline{\phantom{E \rightarrow Cr}}$$
$$Cr$$

and

$$Cr \rightarrow Cs \qquad (2)$$
$$Cr$$
$$\overline{\phantom{Cr \rightarrow Cs}}$$
$$Cs$$

By establishing criteria (Cr), a rule simply guarantees that, upon the receipt of a piece of evidence I, the consequence (Cs) will apply by *modus ponens*; the purpose of legal proceedings being to administer the evidence, that is, to determine whether the evidence is indeed sufficient to establish the criteria or not. This process can involve the weighing of the evidence using probabilistic reasoning (Fenton et al., 2013, 2016; Grabmair & Ashley, 2013; Neil et al., 2019). The consequence (Cs) can be to move to the verification of another criterion (Cr) established by the article; or it can simply be the application of the whole article.

The decision path that can be built out of the aforementioned ontology is presented in figure 2. The path simply states that "if there is evidence for the criterion established by the rule, then the consequence established by the rule should apply"; the decision in this case corresponding to the consequence (Cs) (e.g., applying the rule or not applying the rule, or moving to the next criterion, or not moving to the next criterion). By changing the logical structure of the criteria, complex reasoning patterns (such as alternative and cumulative criteria) can be encoded (Westermann & Benyekhlef, 2023).

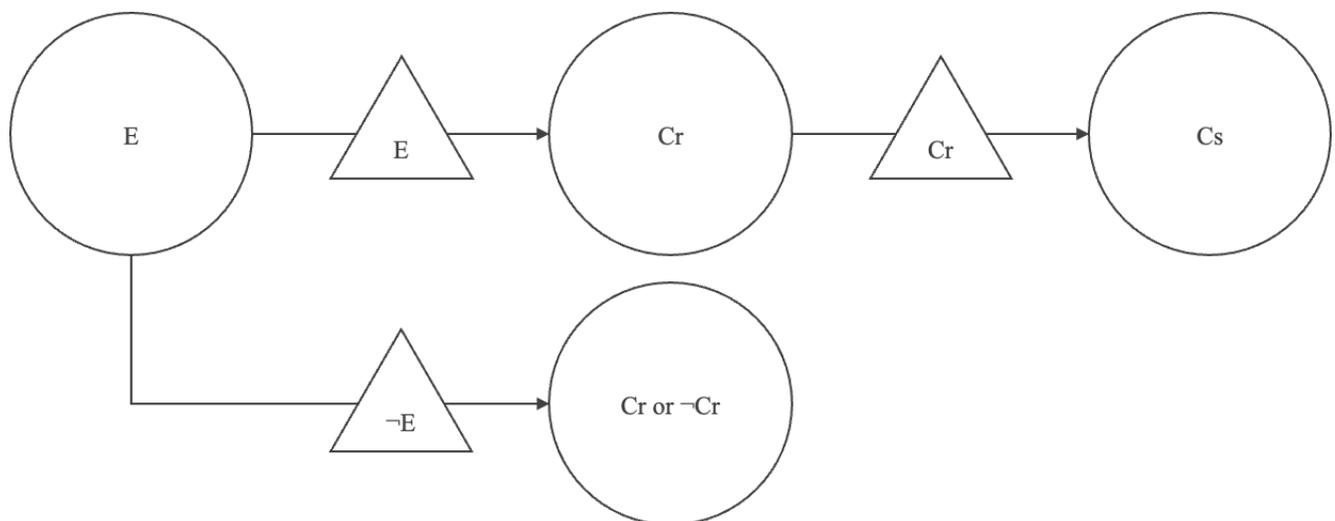

**Figure 2. Flowchart of a logical inference based on a *modus ponens* legal decision path.** When affirming the truth of the implication relation between the evidence I and the criterion (Cr) (E→Cr), if the E applies, then Cr obtains with certainty, and consequently, the consequence (Cs) also obtains with certainty, following a chained application of modus ponens. In turn, if the evidence does not apply (¬E),

then Cr may or may not apply (Cr or ¬Cr). This is so because the negation of E cannot lead to deductively valid inference when affirming E→Cr.

Applying this basic ontology to 38750 b), "an autonomous vehicle may be operated" corresponds to Cs, "on public roads" to Cr1, "for testing purposes" to Cr2, "by a driver who possesses the proper class of license for the type of vehicle being operated" to Cr3, and "if all of the following requirements" to other Crs. Thus, in California, whether an autonomous vehicle should start depends on whether it has evidence that the car is on the public road or not, for testing purposes or not, driven by a driver that has the right license class or not, etc. Decision paths fulfil the explainability requirement for the ETLC system. The decision paths are based on symbolic logic and expert systems. Thus, at every step, it can be determined which criteria were fulfilled, and we can trace the various involved criteria and thus explain to the user why a certain action was taken (e.g. starting the car).

**3.2 – Transforming legislation into the formal representation – Large Language Models**

A key requirement for the ETLC approach described in this paper involves the interoperability of the system. Manually creating decision paths for legislation can be quite tedious, which would limit the ability to use the system across jurisdictions, as a team of legal experts would have to perform the heavy work of encoding legal rules for each new jurisdiction. The ETLC approach described here also envisions the use of automatic methods for transforming legislation into the formal representation described above, i.e. decision paths. The approach to ETLC presented in this section suggests using a Large Language Model (LLM) to convert a legal text – article 38750 b), into the component involved in such an expert system, as per the method used in (Janatian et al., 2023; Westermann & Benyekhlef, 2023).

Multiple such approaches have been tried in the field of computational law. Approaches include the use of language constructs (such as noun and verb phrases) (van Gog R. Sayah K., 2004), morphology (Nakamura et al., 2008), rule-based approaches (Wyner & Peters, 2011) and combination approaches (Dragoni et al., 2016). For the ETLC method described here, we suggest using an approach similar to the one presented in Janatian et al. [5]. Here, the authors used large language models (LLM) to automatically build decision pathways from legislative text. Such models are trained on enormous amounts of text, allowing them to perform sophisticated reasoning in a variety of domains, including the legal domain (Blair-Stanek et al., 2023; Savelka et al., 2023a, 2023b). The authors in [5] used an LLM (GPT-4) to generate a pathway for a decision support tool, which can then be verified and completed by legal experts. The authors selected 40 articles with varying levels of complexity, and provided a prompt to GPT-4 asking it to return the articles in the form of decision paths. Then, they imported the resulting pathway into the "JusticeCreator", a tool to build JusticeBot systems, and compared those pathways to pathways manually created by humans. The results were promising. Overall, for 60% of the articles the researchers saw the automatic and manual pathway as equivalent or preferred the automatically generated pathway. Further, when asked whether the generated pathways could be helpful as the basis for creating a decision pathway, in 90% of the cases the answer was that the generated pathway was either perfect or only required slight adjustments [5].

As applied to our scenario, the query to the LLM, as per the method discussed in [5], involves prompting the LLM with instructions informing it of the desired output format, and providing the traffic rules, such as those in 38750 b). Based on this, the system will create a decision path that can serve as the basis for the formal structure of the ETLC system. Likely, a subject matter expert in the field would have to verify the correctness of the generated pathway. However, hopefully, the generated pathway would be mostly correct, allowing for a rapid and efficient creation of new pathways. The interoperability requirement of the ETLC system would be fulfilled, since it can be rapidly adapted to various jurisdictions, making export of the AIS easier. Importantly, while LLMs are black box systems, since the output of the LLM is an expert system, the explainability requirement would be maintained. While the creation process of the LLM itself may be opaque, the resulting decision path is deterministic and explainable.

**3.3 – Reasoning with a formal representation – Bayesian Networks**

Finally, the generated decision path has to be integrated in the AIS in the course of its actions. This is not always easy, since it can be difficult to assess the applicability of legal criteria in the real world computationally. Legal criteria are often open-ended and vague, which requires human interpretation to apply to specific situations. Expert systems often have trouble representing such fuzzy concepts. Previous solutions include the use of so-called hybrid systems that combine case-based reasoning with rule-based reasoning (Ashley & Brüninghaus, 2009; Rissland & Skalak, 1991), showing the user examples of previous case law summaries (Westermann & Benyekhlef, 2023), and using machine learning (Westermann et al., 2019). Here, we suggest using Bayesian networks to inject Bayesian reasoning under uncertainty as a capability of the AIS, referring to the "C" of the ETLC method.

Bayesian networks are directed acyclic graphs that allow one to represent, compute, and infer efficiently the posterior probability of hypotheses and evidence related by multiple priors and likelihoods; directed acyclic graphs facilitating the identification of conditionally independent nodes in the network (Pearl, 1988). For the purpose of our ETLC method, Bayesian networks are used to enable the AIS that has downloaded a decision path to infer the legally permissible action to be performed, based on observed evidence of fact, which can contain uncertainty. In computational law, Bayesian networks have been traditionally used to model the weighing of evidence in the context of criminal law proceedings, as means of capturing the way one should reason about the relation between a piece of evidence and a verdict, and in particular how to quantify our certainty and uncertainty about pieces of evidence (Neil et al., 2019). More recently, Bayesian networks have been applied to substantial legal reasoning—the reasoning about legal qualification (Constant, 2023). Treating substantial legal reasoning in this Bayesian manner is useful, because Bayesian networks are particularly efficient at solving inference problems that involve sources of uncertainty, and substantial legal reasoning often has to handle the kinds of uncertainty that may stem from seemingly contradictory decisions and from statutory law, which may remain silent on the details of its application.

The application of Bayesian networks to substantial legal reasoning in Constant (Constant, 2023) naturally aligns with the legal ontology of decision paths. This approach demonstrates that Bayesian networks can be used to reconstruct legal syllogisms, which are logical structures used in the law to deduce a legal qualification from facts. This approach thus shows how the ontology of decision paths can be implemented computationally by a Bayesian network. Various kinds of inference algorithms can be used to solve a Bayesian network, the simplest being Bayes rule itself, whereby the posterior probability of a hypothesis (H) conditioned on a piece of evidence I is computed ($P(H|E)$) using the prior probability of the hypothesis ($P(H)$) and the likelihood of the evidence ($P(E|H)$), as well as the joint probability of the evidence and the hypothesis ($P(H,E)$) and the independent probability of the evidence (PI): $P(H|E) = P(H,E)/PI$. Replacing the hypothesis (H) with the criterion (Cr) in the ontology of a decision path, one obtains a hybrid model that combines the probabilistic inference capacities of Bayesian networks with the logical inference capacities of legal decisions paths (figure 3). Provided that the priors are treated as empirical priors, they can be Initialized with probabilities that correspond to the status of the jurisprudence on a given question (e.g., how probable is it that the road is a public road given the features of the road and how those have been treated in case law) (Constant, 2023). In such a model, the evidence I is conditionally related to the criterion (Cr) of Interest to form a likelihood. As illustrated in figure 3, the model of a rule could include 4 likelihood mappings for the conditional probabilities between two possible states of the criterion (e.g., "applies" Cr; not apply ¬Cr) and two possible pieces of evidence (e.g., E1 and E2). Upon the receipt of the evidence, the autonomous intelligent agent (AIS) could infer the posterior probability of Cr and ¬Cr using Bayes rule. Selecting the state with the highest probability (Cr = 0.6316), the AIS could then infer, this time, deductively, the consequence (Cs) that should apply:

$$Cr \rightarrow Cs \tag{3}$$
$$\frac{Cr}{Cs}$$

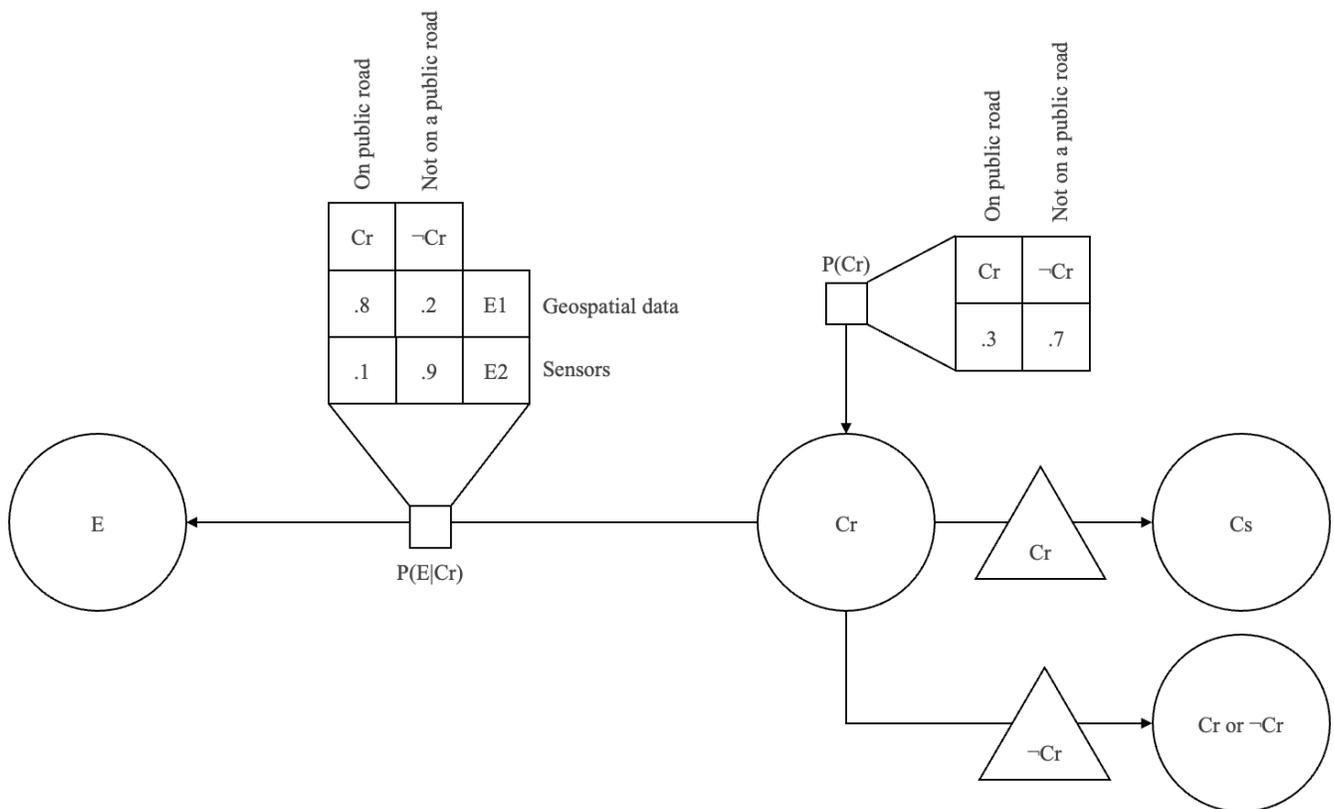

**Figure 3. Depiction of a hybrid Bayesian and logical network for probabilistic and logical inference.** Based on observed evidence, the applicability of the criterion can be inferred, and the consequence logically deduced. P(E|Cr) is the likelihood matrix containing all the likelihoods relating the evidence to the criterion's applicability. The evidence can come from various sources (e.g., geospatial information, E1, or information coming from the car's sensor, E2). The squares indicate the likelihood and prior parameters. The triangles indicate the logical parameter, as per figure 2. Note that instead of using logical parameters, one could use priors and likelihood that contain no uncertainty.

Applying our model to the first criterion of 38750 b), consider a driver that enters an autonomous car and that attempts to start the engine. The AIS controlling the autonomous car then can figure out whether it can operate legally by searching for evidence of the first criterion, namely, whether the car is "on public roads". To do so, the AIS can look at information from a variety of data sources (e.g., geospatial information), and update the parameters of the model in figure 2, namely the prior and the likelihood, accordingly. Assuming that the parameters are empirical parameters, they can be updated using a simple method of adding counts for every information supporting a mapping of interest. For instance, if the sensors capture information that would support the fact that the vehicle is on a public road (e.g., by reading a sign), then the probability of the evidence (e.g., E2) supporting the criterion to the effect that the car is on a public road (Cr) should increase. This could be done by adding a count of +1 to P(E2|Cr), and renormalizing the row corresponding to E2:

P(E2|Cr) = .1     (4)
P(E2|¬Cr) = .9
P(E2|Cr) = .1 + 1 = 1.1
P(E2|Cr) renormalized = .55
P(E2|¬Cr) renormalized = .45

Then, based on the information search, the AIS might find that the evidential relation between the evidence "E" and the criterion (Cr) "on public roads" may contain some uncertainty, for instance, if the evidence comes from a variety of data sources that may conflict due to, for instance, recent changes in municipal regulation. In that case, whether the car is actually on a public or non-public road given the evidence at hand may remain uncertain. However, that uncertainty may be managed by using the empirical prior probability of the criterion being met or not (i.e., the vehicle being on the road or not), based on the history of activities on that road (e.g., if the current road is a dirt road, but that it has been historically used for biking, hiking, or whether it has been frequently maintained by the government). The AIS would iterate through all the available evidence (e.g., E1, E2, etc.), each time updating the prior with the posterior. For instance, after having updated the likelihood based on the sensor information, all likelihoods would be:

P(E1|Cr) renormalized = .8     (5)
P(E1|¬Cr) renormalized = .2
P(E2|Cr) renormalized = .55
P(E2|¬Cr) renormalized = .45

The AIS would then find the posterior for E1:

Cr = (.3*.8)/((.3*.8)+(.7*.2))  = 0.6316     (6)
¬Cr = (.7*.2)/((.3*.8)+(.7*.2)) = 0.3684     (7)

Then the AIS would find the new posterior for E2, using the previous posterior as the current prior:

Cr = (.55*.6316)/((.55*.6316)+(.45*.3684))  = 0.6769     (8)
¬Cr = (.45*.3684)/((.55*.6316)+(.45*.3684)) = 0.3231     (9)

## 4 CONCLUSION

The goal of this paper was to present a method for the extraction, transformation, loading, and computation (ETLC) of legal text. ETLC of legal text faces two main challenges; the challenge of interoperability (i.e., deployability across different jurisdictions, which have different laws) and the challenge of explainability/auditability (i.e., explainability and auditability of AIS decision making). The motivation for the proposed model was the realization that current AIS regulation does not regulate AIS *per se* (but rather the developers). We suggested that, in order to achieve an effective regulation of AIS, two paths are available. We argued that the first path to AIS regulation was that of "teaching" AIS how to behave, in a way that conforms to existing regulation applying to them indirectly via the regulation of AIS driven devices, such as autonomous vehicles and IoT devices. The ETLC method developed in this paper functions as such a method of teaching AIS existing regulation, so as to allow them to behave autonomously and legally; what we called legal autonomy.

We argued that any ETLC system for AIS should be interoperable, i.e., should be domain general with respect to the rule being learned; and that they should be explainable, i.e., that they should enable human-interpretable descriptions of the reasoning process underlying the decision made by the AIS. The ETLC method presented in this paper tackles the challenges of interoperability by leveraging LLMs and a legal ontology that should apply to any rule in any jurisdiction: the ontology of evidence, criteria, and consequence of legal decision paths. The ETLC method proposed here also tackles the challenge of explainability by using the human interpretable logic of legal decision paths, combined with the computational capability of Bayesian networks. The probabilistic reasoning in Bayesian networks is fully interpretable, and provides an explanation of the decision partly based on concepts such as "probability of application of criteria" and "weighting of evidence" that are familiar to most people working with probabilistic and inferential systems. Inquiring about the reasoning process of an agent that, for example, refused to start the engine of the autonomous car under its control, one could find that "the agent did not start the car because based on evidence (e.g., sensor information and geospatial data), there was a 63.16% chance that the car was on a public road." As shown above, this reasoning is fully transparent from the point of view of our ETLC methods. Of course, probabilistic reasoning mostly makes sense for civil law situations that use a "balance of probability" standard of proof, where the evidence for the fulfilment of a criterion must "make it more probable that improbable" that a criterion applies, and therefore that its consequence ensues. Whether it would be socially acceptable to allow agents to make decisions that are not supported by evidence at 100% is an open question (e.g., the agent should allow the driver to start the car unless the agent is 100% certain that it is not on a public road, or vice versa). That being said, as far as legal matters are concerned, decisions based on a concept of balance of probability given available evidence should be sufficient to justify and explain AIS decisions.

Of note, the use of Large Language Models (LLMs) was instrumental to the method described in this paper. Building decision paths out of legal texts involves a cumbersome coding process—the process of manually determining what are the evidence, criteria and consequences of a rule. If we had to go through such a coding process each time that an amendment to a law relevant to AIS' behaviour was made or that such a law was adopted, it would be impossible to keep AIS updated with the currently binding law. Going back to our manufacturer earlier, let's imagine that that manufacturer loses its case and the media gets wind that an autonomous device was responsible for causing serious harm to a natural person. The Australian, French and American legislators then react and adjust their regulation to impose more stringent requirements on the operations of autonomous devices. Will the manufacturer have to send a software update to append to the existing software the change in the regulation? Will the autonomous agents driving the decision of the manufacturers' autonomous cars have to be retrained? Not with the ETLC method proposed in this paper, thanks to the combination of LLMs, decision paths, and Bayesian networks.

Beyond the compensation of damages, our manufacturer may face additional troubles if it is not in a position to explain the decision having led to the accident; hence the importance of AI explainability stressed in section 2. The inability to understand or predict how AI systems make decisions can lead to a general erosion of trust among users and the broader public. When people cannot ascertain the rationale behind an AI's decision, especially in critical applications affecting health, safety, or privacy, it undermines confidence in the technology. This scepticism may hinder the adoption of AI solutions and stifle ethical innovation (Kożuch & Sienkiewicz-Małyjurek, 2022; Toreini et al., 2019). Another important side of the problem of explainability is the risk of overreliance on AI systems, where individuals and institutions place undue trust in the infallibility of algorithmic decisions from black boxes that for different reasons may be trusted despite their lack of explainability (e.g., because of the reputation of the company producing the black box AIS) (Pavlidis, n.d.; von Eschenbach, 2021). This overtrust can lead to complacency, where the outputs of AI systems are accepted without question, potentially overlooking biases, inaccuracies, or errors embedded within the AI's decision-making process (Casper et al., 2024; Harbarth et al., 2024).

In this paper, we have focussed on the "legal" explainability of AIS decisions. However, the ability to explain and justify decisions made by AI systems transcends legal exigencies and touches the core of ethical responsibility and accountability. If a decision made by an AI system leads to harm or death, the moral responsibility of that outcome necessitates that the decision-making process be scrutinized and understood. The principle that decisions must not only be made but also justified is deeply rooted in ethical philosophy. This principle is especially relevant in the context of AI, where decisions need to be explainable to be ethically and legally justified. If an AI system's decision cannot be explained, it raises serious questions about the system's ethical deployment. In life-critical applications, the stakes of decisions are so high that unexplainable decisions are ethically untenable. In the context of life-critical and weapon systems, there must be a clear chain of accountability for decisions made by AI (Ferl, 2024). This accountability is only possible if decisions can be explained and traced back to their underlying logic and the data on which they were based. Without explainability, assigning responsibility for the outcomes of AI decisions becomes fraught with difficulty, undermining legal and ethical accountability frameworks. The social license to operate advanced AI systems in sensitive and high-stakes domains is contingent upon public trust (Zhang et al., 2024). This trust should be predicated on the assurance that AI systems operate within ethical boundaries and that their decisions are subject to oversight and justification. Explainability is a cornerstone of building and maintaining this trust, as it allows the public and regulatory bodies to understand and evaluate the decision-making processes of AI systems.

Tošić, P. T. (2016). Understanding autonomous agents: A cybernetics and systems science perspective. *2016 Future Technologies Conference (FTC)*, 121–129.

Turner, J. (n.d.). *Robot Rules*. Springer International Publishing.

van Gog R. Sayah K., van E. T. M. (2004). A case study on automated norm extraction. In Jurix (Ed.), *Legal Knowledge and Information Systems* (pp. 49–58). IOS Press.

von Eschenbach, W. J. (2021). Transparency and the Black Box Problem: Why We Do Not Trust AI. *Philosophy & Technology*, *34*(4), 1607–1622.

Walker, V. R. (2006). A default-logic paradigm for legal fact-finding. *Jurimetrics*, *47*(193).

Walker, V. R., Okpara, N., Hemendinger, A., & Ahmed, T. (2017). *Semantic types for decomposing evidence assessment in decisions on veterans' disability claims for PTSD*. https://scholarlycommons.law.hofstra.edu/faculty_scholarship/1134/

Westermann, H. (2023). *Using artificial intelligence to increase access to justice*. https://doi.org/1866/32168

Westermann, H., & Benyekhlef, K. (2023). Justicebot: A methodology for building augmented intelligence tools for laypeople to increase access to justice. *ICAIL*, 351–360.

Westermann, H., Walker, V. R., Ashley, K. D., & Benyekhlef, K. (2019). Using Factors to Predict and Analyze Landlord-Tenant Decisions to Increase Access to Justice. *Proceedings of the Seventeenth International Conference on Artificial Intelligence and Law*, 133–142.

Westhofen, L., Stierand, I., Becker, J., Möhlmann, E., & Hagemann, W. (12 2022). Towards a Congruent Interpretation of Traffic Rules for Automated Driving - Experiences and Challenges. *Proceedings of the International Workshop on Methodologies for Translating Legal Norms into Formal Representations (LN2FR 2022) in Association with the 35th International Conference on Legal Knowledge and Information Systems (JURIX 2022)*.

Wilson, B., & Cali, S. (2016). Smarter cities, smarter regulations: a case for the algorithmic regulation of platform-based sharing economy firms. *UMKC Law Review*, *85*, 845.

Wyner, A., & Peters, W. (2011). On rule extraction from regulations. In *Legal knowledge and information systems* (pp. 113–122). IOS Press.

Zhang, W., Shanmugam, S., Allen, J. G., Anon, W., & Xu, O. (2024). Auditing AI for whom? A community-centric approach to rebuilding public trust in Singapore. *SSRN Electronic Journal*. https://doi.org/10.2139/ssrn.4698703

Zurek, T., Kwik, J., & van Engers, T. (2023). Model of a military autonomous device following International Humanitarian Law. *Ethics and Information Technology*, *25*(1), 15.
17